# Deployment Dynamics and Optimization of Novel Space Antenna Deployable Mechanism

Mamoon Aamir[a], Mariyam Sattar[a*1], Naveed Ur Rehman Junejo[b*2], Aqsa Zafar Abbasi[c]

[a]Department of Aeronautics and Astronautics, Institute of Space Technology, Islamabad, Pakistan
[b]The University of Lahore, Lahore Campus, Pakistan.
[c]Department of Applied Mathematics and Statistics, Institute of Space Technology, Islamabad, Pakistan

[*1]Corresponding Author 1: mariya98975@gmail.com
[*2]Corresponding Author 2: naveed.rehman@dce.uol.edu.pk

**Abstract**

Given the increasing need for large aperture antennas in space missions, the difficulty of fitting such structures into small launch vehicles has prompted the design of de- ployable antenna systems. The thesis introduces a new Triple Scissors Deployable Truss Mechanism (TSDTM) for space antenna missions. The new mechanism is to be stowed during launch and efficiently deploy in orbit, offering maximum aperture size while taking up minimal launch volume. The thesis covers the entire design pro- cess from geometric modeling, kinematic analysis with screw theory and Newtonian approaches, dynamic analysis by eigenvalue and simulation methods, and verification with SolidWorks. In addition, optimization routines were coded based on Support Vector Machines for material choice in LEO environments and Neural Networks for geometric setup. The TSDTM presented has enhanced structural dynamics with good comparison between simulation and analytical predictions. The structure op- timized proved highly accurate, with a deviation of just 1.94% between machine learning-predicted and simulated natural frequencies, demonstrating the potential of incorporating AI-based methods in space structural design.

**Keywords-** DoF; Kinematic Analysis; Triple scissors link deployable antenna mechanism; Virtual experiments; Storage ratio

**Introduction**

Deployable mechanism is a type of mechanism which can be changed from a folded compact form to an expectant deployed state, and can be a full stable structure with capacity of load adaptation [1, 2]. Because of the excellent performance in space applications, deployable mechanisms are extensively applied to construct large space structures, e.g., deployable mast, deployable antenna, etc., and are of great importance in space missions like earth observation, telecommunications, scientific researches etc. Due to the high storability and light weight, the deployable antenna with flexible cable net is one of the most de-sired antennas in aerospace applications [3, 4]. Numerous efforts, such as innovative design concepts, analysis methods and experiments, were devoted to improve the deployable antennas technology. Large deployable antennas have been built for space missions with different structural schemes, however, most of them can be classified as radial structures (e.g. Lockheed's Wrap- Rib antenna, Harris' Rigid-Rib and ESA's MBB antenna Hinged- Rib), modular structures (e.g. JAXA's ETSVIII, Tashkent's KRT10, OKB-MEI's TKSA-6) and peripheral truss (e.g. Northrop-Grumman's AstroMesh, Harris' hoop-truss and hoop-column, GTU's MIR reflector experiment, ESA's LDA) [5, 6].

Of these, peripheral ring antennas having the benefit of self-synchronization, high thermo-elastic stability, and deployment dependability have been studied for the last few years. Since the early 1980s Soviet space programs originated at GTU, and studies in some pantograph ring concepts and related technologies have been conducted [7, 8]. In 1999 a 5.5-m peripheral pantograph structure with radial tensed membrane ribs was flown and deployed on the MIR station. Pantographs and derived linkages have been studied in great detail at GTU, TUM, and other research groups [9]. Conversely, double ring structure, like Cambridge's Deployable Mesh Reflector, is made up of two concentric and peripheral pantograph rings of unequal heights connected radially by a third set of pairs of pantographs, despite the seeming complication, the whole mechanism has single mobility and high stiffness and accuracy [10]. Chinese scholars have put forward two categories of double-ring deployable truss ideas based on parallelogram mechanism, structural rigidity is confirmed through prototype testings [11]. Astro Aerospace Company has designed one type of AstroMesh deployable antenna composed of two symmetrical parabolic cable nets, one metal mesh reflector surface, and one deployable ring mechanism, they own more than 15 years of uninterrupted developing history [12, 13]. JAXA also created two truss antennas of 19-meter diameter, which apply to the satellite communication service with engineering test satellite ETS-VIII launched in December 2006. The antenna was sustained by 14 hexagonal truss modules of diameter 4.9 meters, and each module consists of a mesh surface, cable network, and deployable structures made up of six basic deployable units [14]. A research group headed by L. Datashvili proposed the double pantograph-based peripheral ring [15], which meets the requirement of much less mass and smaller folded volume with stability and deployment reliability. Besides the above-discussed design issues, our research team in HIT has also made great contributions in the field of large deployable mechanisms, such as the synthesis of deployable mechanisms [16], mobility analysis [17], optimization design [18], and cable net form-finding [19]. This paper presents our recent effort in a new concept design for a large mesh deployable antenna.

In response to the growing requirement for satellite communications services and Earth observation missions, we design a ring-based structure made of multi-deployable modules. In the present work, a new Triple Scissor Deployable Truss Mechanism (TSDTM) is presented as a viable option for deployable structures in space missions. The TSDTM is made up of modularly linked scissor-shaped units in a truss configuration that can be folded and deployed efficiently with high structural integrity. Compared to traditional deployable trusses, the mechanism has better compactness in the deployed or stowed condition and stiffness in the fully deployed condition, due to which it is highly suitable for large antennas. The benefits of the TSDTM are its high deployment ratio, structural efficiency, and versatility in a wide range of space-borne applications. Utilizing the kinematic characteristics of the scissor mechanism, the designed structure possesses geometric stability in deployment with the lowest mechanical complexity. Moreover, the modularity of the design enables scalable realization, qualifying it for applications that entail the use of large-aperture antennas in satellite communications and remote sensing.

This work is based on previous achievements in deployable mechanism synthesis and optimization. It aims to advance the development of highly efficient and dependable space structures. Future research will focus on refining structural optimization, incorporating actuation mechanisms, and experimental validations for real-world performance under simulated space conditions. An exhaustive study was performed to verify the Triple Scissor Deployable Truss Mechanism (TSDTM). The mechanism was initially designed in SolidWorks, followed by analytical kinematic analysis to compare its motion properties. To authenticate these results, a multi-body simulation using MSC ADAMS was done with the guarantee of correct kinematic behavior. Then, dynamic analysis with the Lagrangian method is performed to find the natural frequencies and structural stability. To further confirm these results, finite element simulations in ANSYS were performed, which verified the analytical predictions. This combined approach guarantees the reliability and efficiency of TSDTM for space missions, especially in large-aperture deployable antennas.

It is crucial to estimate the natural frequency of the TSDTM in order to keep its stability and dynamic performance during deployment and operation in space. The natural frequencies were determined using analytical derivations and finite element analysis was performed to validate the results. The results showed that the structure remains free from resonant vibrations. A well-optimized range for natural frequency is needed to eliminate structural resonance. The findings verified that TSDTM has a high natural frequency 0.1182, which is beneficial for the structural design for space. Further refinement and development will be targeted to the structural damping characteristics to make it more rigid during real work conditions.

**Problem Statement**

Despite their critical role in satellite communications and Earth observation, large-aperture space antennas face significant challenges related to compactness, deployment efficiency, and structural stability. Conventional deployable mechanisms, such as single-ring and double-ring trusses, often exhibit limitations in mechanical reliability and storage efficiency. This research introduces the Triple Scissor Deployable Truss Mechanism (TSDTM), a modular structure designed to enhance compactness in the stowed configuration and structural rigidity in the fully deployed state. The study employs kinematic analysis, dynamic simulations using MSC ADAMS and ANSYS, and finite element analysis to evaluate TSDTM's performance. The findings demonstrate its superiority over conventional mechanisms in terms of deployment time, natural frequency, and storage efficiency.

**Proposed Design for Single Deployable Unit:**

The basic building block of the modular mechanism was initially designed to be a single scissors link. The basic modular units in the antenna are interconnected to form a circular deployable mechanism, which is developed in a methodical manner to ensure a clear progression from a simple configuration to the final design. Fig. 1 shows the step-by-step development of modules from basic scissors links and their integration into the finalized basic deployable unit.

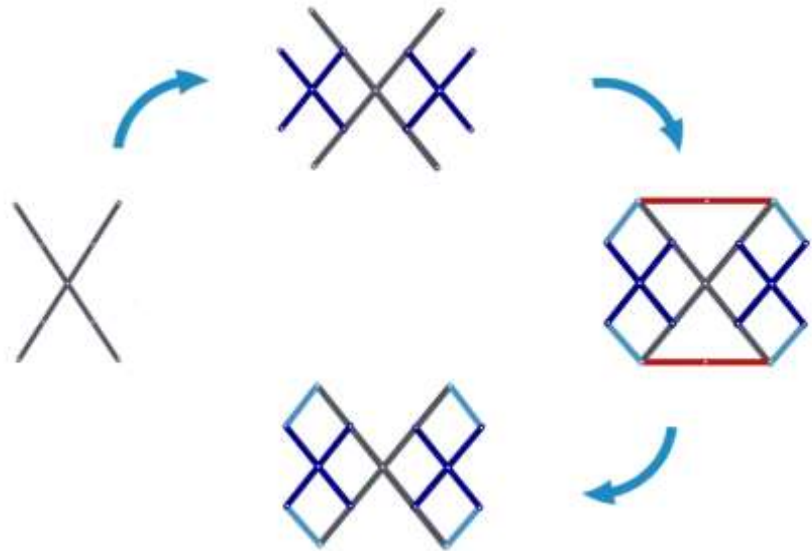

**Fig. 1** Step-by-step development of modules from basic scissors links

Then two more scissor links were symmetrically added on either side of the first link to increase the overall span and create the core structure, to improve structural support and stability, supporting links were added between the scissor units. These links were essential for keeping the expanding structure aligned and rigid. Finally, four horizontal links were incorporated to complete the single unit.

The triple scissors links assembly's top and bottom horizontal links fixed the modular unit's stretched length and guaranteed even load distribution. This methodical process made it possible for an orderly assembly, guaranteeing that every phase added to the antenna's overall stability and performance. The top and bottom links are retrained at an angle of 50°, whereas the fully deployed modular unit is constrained at an angle of 80° between the scissors links. The high storage ratio that this innovative unit with triple scissors links mechanism is designed to provide makes it ideal for large-aperture space antennas with strong structural integrity. Ample surface area for efficient signal capture and transmission is provided by the 25m aperture. The TSDTM's detailed stowed structure is shown in Fig. 2. The angle between the scissors and the top and bottom links is 12.54° when the device is fully stowed.

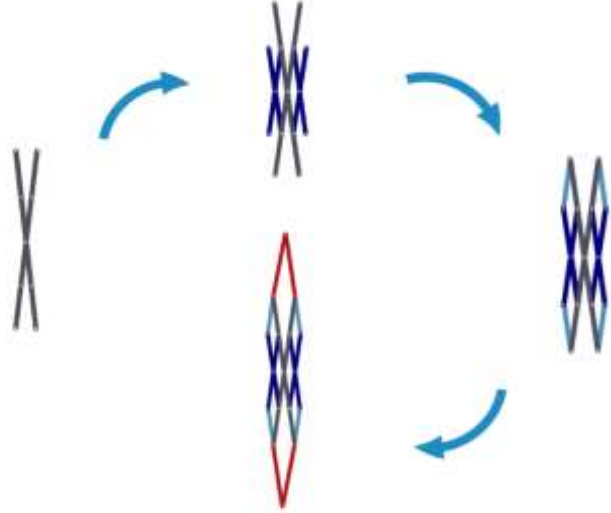

**Fig. 2** Detailed stowed structure

**Geometric Design of Single Deployable Unit:**

Fig. 6 shows that the basic deployable unit contains four horizontal links and ten diagonal links. This modular assembly was configured for an antenna of 25m aperture. A detailed comparative analysis of deformation patterns and storage ratios is performed for a 25m aperture antenna formed from 12, 18 and 24 deployable modular units.

Due to its exceptional structural integrity and performance, the space antenna truss assembly consisting of 12 deployable modular units, has been chosen for more study. As a result, 6.47 meters is the estimated stretched length of the base module for a 12-unit antenna assembly. The angle between the two-diagonal scissors links, designated $L_1$ and $L_2$, was limited to 80° for the fully deployed structure in order to attain the optimal deployment. For the triple scissors link modular unit, the exact dimensions of each link were methodically calculated using geometry and trigonometric connections.

$$\tan(40°) = \frac{L_3}{2.545} = 2.14m \tag{1}$$

As the unit is symmetrical; the lengths $L_3$, $L_4$, $L_5$, and $L_6$ are equal. To calculate the lengths of diagonal links $L_1$ and $L_2$, we consider a right triangle and set the link $L_1$ as hypotenuse. The angle between links $L_1$ and $L_3$ is 50°. The mathematical formulation given in equation (2) is applied to calculate the length of links $L_1$ and $L_2$.

$$\sin(50°) = \frac{5.09}{L_1} = 6.645m = L_2 \tag{2}$$

The length of diagonal links $L_7$, $L_8$, $L_9$ and $L_{10}$ were half of length for link $L_1$, equation (3).

$$L_7 = L_8 = L_9 = L_{10} = 3.323 m \qquad (3)$$

Similarly, the length of diagonal links $L_{11}$, $L_{12}$, $L_{13}$ and $L_{14}$ were half of the link $L_7$.

$$L_{11} = L_{12} = L_{13} = L_{14} = 1.662 m \qquad (4)$$

The dimensions calculated for 14 different links that constitute the triple scissors links deployable modular unit are given in Table 1.

**Table 1.** Length (L in meters) for all links of the modular unit

| Link | $L_1$ | $L_2$ | $L_3$ | $L_4$ | $L_5$ | $L_6$ | $L_7$ | $L_8$ | $L_9$ | $L_{10}$ | $L_{11}$ | $L_{12}$ | $L_{13}$ | $L_{14}$ |
|---|---|---|---|---|---|---|---|---|---|---|---|---|---|---|
| L (m) | 6.64 | 6.64 | 2.14 | 2.14 | 2.14 | 2.14 | 3.32 | 3.32 | 3.32 | 3.32 | 1.66 | 1.66 | 1.66 | 1.66 |

Stretch length, deploy/stow height, deploy/stow diameter, deploy/stow volume, and storage ratio calculations are carried out for antennas with 12, 18, and 24 modular modules while fixing the aperture at 25 meters. Table 2 illustrates how the structure can be set up with different numbers of modular pieces based on the necessary aperture size. Deformation evaluations of the 18 and 24 unit modular assemblies revealed substantial levels of structural instability, despite the fact that they provided a good compromise between efficient deployment and compact stowage. The antenna assembly, which consists of 12 modular parts, maintains a consistent deployed diameter of 25m while offering a greater deployed height of 5.122m. Additionally, it exhibits a minimal deformation of 0.01048mm without sacrificing structural integrity, which makes it more appropriate for the demands of our particular task. In the proposed design, there are two independent variables. The diameter of the antenna, which is determined by its intended function, is the first consideration. The goal of this study is to create a large deployable antenna with a diameter of 20 to 30 meters. Consequently, a deployable antenna with a diameter of 25 meters has been chosen for this experiment. The height of each link, the length of a fully deployed unit, the length of moveable slots, and the diagonals are all determined by the second independent variable, or the number of modular units. Using SolidWorks, a full-scale assembly with 12 modular pieces and a 25-meter aperture was created. Each part, such as the center linkages and interconnected scissor mechanisms, was depicted in depth by this 3D assembly.

**Table 2.** Evaluation of design parameters for 12, 18 and 24 modular unit antenna assembly

| Units | 12 | 18 | 24 |
|---|---|---|---|
| **Stretched Length (m)** | 6.470 | 4.34 | 3.26 |
| **Deployed Height (m)** | 5.09 | 3.436 | 2.581 |
| **Stowed Height (m)** | 11.010 | 7.386 | 5.548 |
| **Deployed Diameter (m)** | 25 | 25 | 25 |
| **Stowed Diameter (m)** | 3.246 | 2.176 | 1.634 |
| **Deployed Volume (m³)** | 2400.584 | 724.61 | 307.109 |
| **Stowed Volume (m³)** | 86.979 | 26.211 | 11.114 |
| **Storage Ratio (Diameter)** | 7.702 | 11.5 | 15.3 |
| **Storage Ratio (Height)** | 0.465 | 0.465 | 0.465 |
| **Storage Ratio (Volume)** | 27.6 | 27.6 | 27.6 |

For link length ($L$), angle ($\theta$), overall width ($W$), and height ($H$), the stowed and deployed angles can be computed. The deployed and stowed states of modular mechanisms are calculated using the cosine laws, as shown in equations (5) and (6), respectively.

$$W = L\sqrt{2(1 + \cos(\theta_1))} \quad (5)$$

$$H = L\sqrt{2(1 + \cos(\theta_2))} \quad (6)$$

Where $\theta_1$ and $\theta_2$ in the above equations are the deployed and stowed angles respectively. Substituting $W = 5.09m$ and $L = 3.32m$ in equation (1) verifies $\theta_1 = 79.90° \approx 80°$.

A table 5 and table 6 illustrates key measures of a deployable space truss mechanism demonstrating how different structural dimensions vary with increasing aperture size. Consistent scaling appears to be achieved; that is, larger apertures correspond to proportionally larger values of stretched length, deployed height, and deployed volume. This demonstrates that the mechanism has a modularity of structure design where uniform relationships between all units are achieved with increasing total size.

A very critical part of this mechanism is that its stowed configurations are so different from its deployed configurations. This clearly reflects how compact the folded configuration can be to accommodate for storage and transport, because it's considerably shorter and thinner compared to the diameter of its deployment configuration. Stowed volume also differs, since it's quite a bit smaller than deployed volume, an aspect important for the success of missions in space: the use of payload capacity efficiently. Further confirming the design efficiency, there are the storage ratios. The latter is around 7.7 for diameter, meaning that the reduction factor is relatively uniform when the truss is stowed. For height, the storage ratio is constantly at 0.465, meaning that when stowed, it is always at 46.5% of deployed height. Most remarkably, for volume, at all aperture sizes, the storage ratio is at 27.6, representing that the deployed truss space in space would occupy 27.6 times more volume in its stowed form. In fact, over such a scale of aperture size, such consistent results re-emphasize the scalability and the reliability of this truss mechanism. In general, the table indicates that a deployable space truss mechanism was developed with the focus on having storage efficiency, predictable scalability, and optimized deployment. Uniform ratios in storage assert that it's a good engineering system, having the same adaptability to be achieved for applications in space structures, making the mechanism highly suited for space structures where compact storage and large dimension in deployment are critical.

**Table 5.** Evaluation of design parameters for with links antenna assembly for different apertures

| Aperture | 6m | 13m | 15m | 25m | 28m | 30m |
|---|---|---|---|---|---|---|
| **Stretched Length** | 1.553 | 3.365 | 3.882 | 6.470 | 7.247 | 7.765 |
| **Deployed Height** | 1.230 | 2.664 | 3.073 | 5.122 | 5.738 | 6.148 |
| **Stowed Height** | 2.643 | 5.726 | 6.606 | 11.010 | 12.332 | 13.214 |
| **Deployed Diameter** | 6 | 13 | 15 | 25 | 28 | 30 |
| **Stowed Diameter** | 0.779 | 1.688 | 1.947 | 3.246 | 3.635 | 3.895 |
| **Deployed Volume** | 33.214 | 337.733 | 518.493 | 2400.584 | 3374.006 | 4150.358 |
| **Stowed Volume** | 1.203 | 12.235 | 18.788 | 86.979 | 122.224 | 150.369 |
| **Storage Ratio (Diameter)** | 7.702 | 7.702 | 7.704 | 7.702 | 7.703 | 7.702 |
| **Storage Ratio (Height)** | 0.465 | 0.465 | 0.465 | 0.465 | 0.465 | 0.465 |
| **Storage Ratio (Volume)** | 27.6 | 27.6 | 27.6 | 27.6 | 27.6 | 27.6 |

**Table 6.** Evaluation of design parameters for without links antenna assembly for different apertures

| Aperture | 6m | 13m | 15m | 25m | 28m | 30m |
|---|---|---|---|---|---|---|
| **Stretched Length** | 1.553 | 3.365 | 3.882 | 6.470 | 7.247 | 7.765 |
| **Deployed Height** | 1.230 | 2.664 | 3.073 | 5.122 | 5.738 | 6.148 |
| **Stowed Height** | 1.607 | 3.483 | 4.018 | 6.697 | 7.501 | 8.037 |
| **Deployed Diameter** | 6 | 13 | 15 | 25 | 28 | 30 |
| **Stowed Diameter** | 0.779 | 1.688 | 1.947 | 3.246 | 3.635 | 3.895 |

| Deployed Volume | 33.214 | 337.733 | 518.493 | 2400.584 | 3374.006 | 4150.358 |
|---|---|---|---|---|---|---|
| Stowed Volume | 0.732 | 7.443 | 11.427 | 52.906 | 74.343 | 91.457 |
| Storage Ratio (Diameter) | 7.702 | 7.702 | 7.704 | 7.702 | 7.703 | 7.702 |
| Storage Ratio (Height) | 0.765 | 0.765 | 0.765 | 0.765 | 0.765 | 0.765 |
| Storage Ratio (Volume) | 45.4 | 45.4 | 45.4 | 45.4 | 45.4 | 45.4 |

Below is a summary of the measurement of a deployable space truss mechanism in the collapsed or without links configuration for comparison with the earlier configuration that included links. The data follows a scaling pattern where larger apertures translate to increases in stretched length, deployed height, and eventually a proportional rise in the deployed volume. Stowage dimensions along with storage ratio significantly differ pointing towards the effective role of a removed link about system compactness. In comparison to the linked configuration, the stowed height in the without links version is significantly lower, resulting in a higher storage ratio for height (0.765 vs. 0.465). This means that the without links structure is more compact in the stowed state. Furthermore, the stowed volume is much less, which leads to a remarkably increased storage ratio for volume (45.4 vs. 27.6), showing superior volumetric efficiency in the without links configuration. These notwithstanding, the storage ratio for diameter (~7.7) is the same for both configurations, indicating that folding behavior in the radial direction is not affected by the presence or absence of links. In general, the without links deployable space truss mechanism has superior storage efficiency, especially in terms of height and volume, thus making it a more compact and efficient alternative for space applications where minimizing stowed dimensions is a critical design consideration.

**Deployment Simulation SOLIDWORKS:**

The deployment sequence of the 12-module deployable structure's interconnected units, where each unit is connected to the next, is depicted in Fig. 3 For TSDTM, the entire cycle takes 102 seconds, and the stowed to fully deployed condition is reached in 53 seconds. Table 3 presents the comparative study for the suggested 25m aperture TSDTM.

**Table 3.** Comparative analysis of deployment process time of deployable antenna in seconds

| Antenna Mechanism | Aperture (m) | No. of units | Deployment Process time (sec) | | Complete cycle (sec) |
|---|---|---|---|---|---|
| | | | Intermediate | Complete deployed | |
| **Triple scissor** | 25 | 12 | 26 | 53 | 102 |
| **Single ring truss** | 25 | 24 | 50 | 140 | -- |
| **Double ring truss** | 5 | 18 | 60 | 102 | -- |

It can be observed that the deployment of proposed deployable mechanism is more efficient than the single ring and double ring truss antenna that required 140s and 102s, respectively, to achieve fully deployed state.

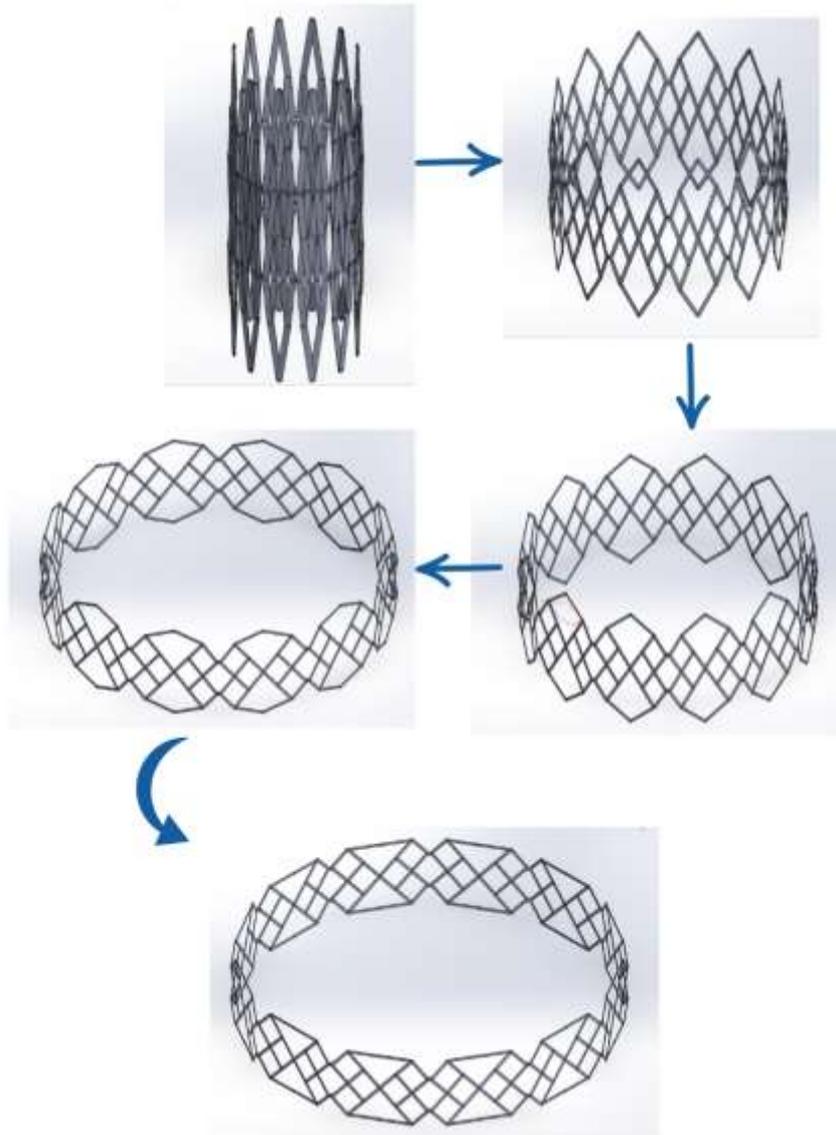

**Fig. 3** Cycle of operation for TSDTM assembly i.e., stowed – intermediate – fully deployed – internmediate – stowed state of 12 modular units antenna

Fig. 4 provides a zoomed-in view of the joint, highlighting its design and role in enabling smooth deployment and structural cohesion.

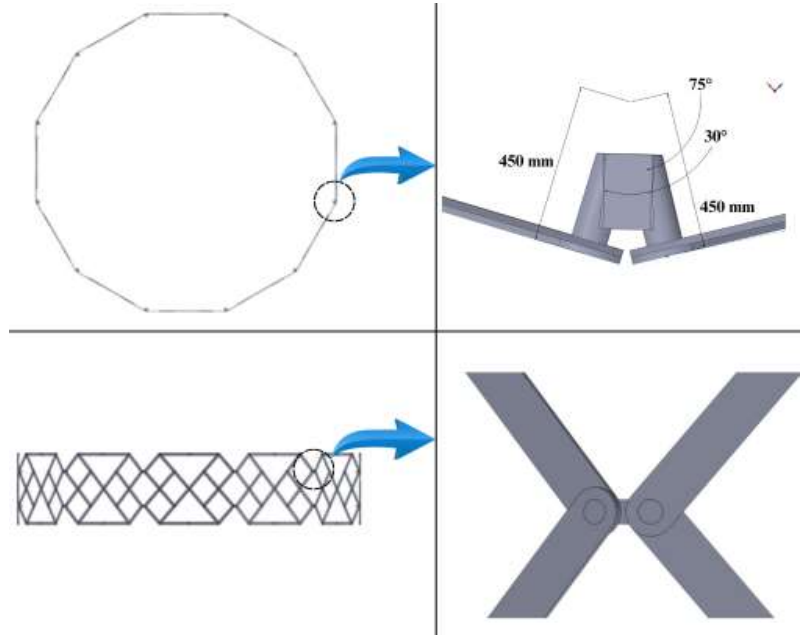

**Fig. 4** Zoomed – in view of the joints involved in smooth deployment of TSDTM

**Mobility Analysis**

The novel deployable truss mechanism under investigation is designed to achieve a single degree of freedom (DoF) to facilitate controlled deployment and stowage. The mobility of the structure is determined using Gruebler's equation for planar mechanisms:

$$M = 3(n - 1) - 2j_p - j_h$$

where M represents the degree of freedom, n is the total number of links, $j_p$ denotes the total number of primary joints (lower pairs), and $j_h$ accounts for the higher-order joints.

For the proposed truss structure, the following parameters have been determined: $n = 18, j_p = 26, j_h = 0$.

Substituting these values into Gruebler's equation:

$$M = 3(18 - 1) - 2(26) - 0$$
$$M = 51 - 52$$
$$M = 1$$

This result confirms that the mechanism possesses a single degree of freedom, ensuring synchronized deployment through a single actuation input. Each modular segment of the truss operates independently in terms of kinematic motion, with overall synchronization facilitated via deployment cables. Consequently, the entire assembly maintains a DoF of 1, validating its suitability for deployable aerospace applications.

**Mesh and Mesh convergence study**

Fig. 5 shows the meshed geometry of the 12 modular units and 25m aperture TSDTM. The precise mesh of the deployable antenna mechanism ensures accuracy of results in the virtual experiments. The meshing performed in SolidWorks employed tetrahedrons to discretize the antenna assembly. The initial mesh consisted of 0.51 million with element quality of 0.5619 that followed subsequent mesh refinements. For reliable analysis, the quality of the mesh was assessed considering the factors such as the skewness of the elements and their aspect ratio.

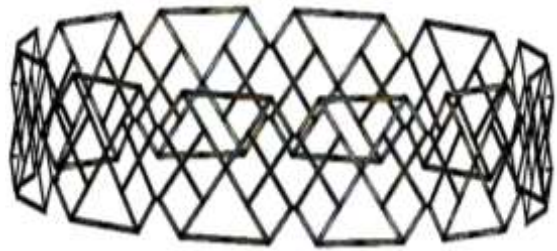

**Fig. 5** Meshed geometry for 12 modular units TSDTM assembly

The results of the mesh convergence study and the change in the orthogonal quality after subsequent mesh refinements are shown in Fig. 6 and Table 4 respectively. This analysis reveals that further enhancement of the orthogonal quality to reduce skewness results in insignificant variation in the results but increases the time for computation. To find the ideal mesh density, indicating mesh-independent outcomes, a total of nine iterations were performed. The final mesh consisting of 1.56 million tetrahedral-shaped elements provided the best balance between computation speed and solution accuracy.

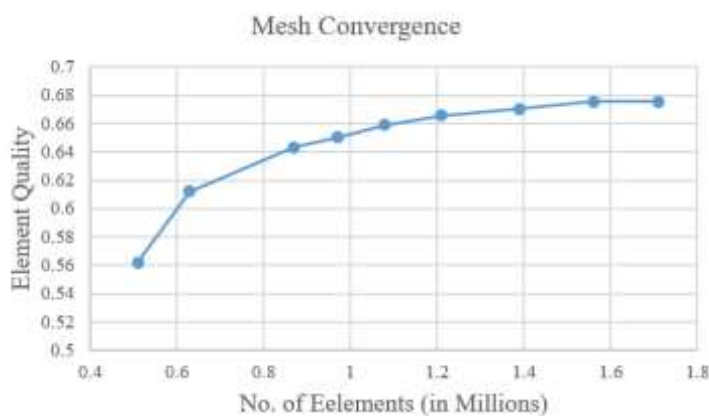

**Table 4.** Mesh Convergence Study

| Sr. # | No. of Elements | Element Quality |
|---|---|---|
| 1 | 0.51 | 0.5619 |
| 2 | 0.63 | 0.6121 |
| 3 | 0.87 | 0.6433 |

**Fig. 6** Mesh convergence analysis based on number of elements against their element quality

| 4 | 0.97 | 0.6501 |
|---|------|--------|
| 5 | 1.08 | 0.6591 |
| 6 | 1.21 | 0.6656 |
| 7 | 1.39 | 0.6705 |
| 8 | 1.56 | 0.6754 |
| 9 | 1.71 | 0.6754 |

**Kinematics Analysis**

In the mechanism two modular units exhibit a motion independent from each other. The motion of the mechanism depends on the speed of the deployment rope monitored by the speed of the electric motor. Therefore, the kinematic analysis is performed on a single unit model. An independent study of a unit module is sufficient and holds for each unit of the structure. The kinematic analysis is performed to determine, (a) The translational and angular velocity of parts (b) The translational and angular acceleration of parts. The deployment process is very slow in order to keep the velocities, accelerations and induced forces at joints to a minimum. The value for slider movement was selected to be 0.1 m/s after running a number of simulations.

**Assumptions & System Description**

- Point "$O$" is the only fixed point in the structure.
- Other points $(A, B, C, ..., Q)$ have different constraints based on the system's configuration.
- The structure follows Newtonian mechanics for kinematics.
- Velocity & acceleration relations are derived based on rigid-body assumptions.

**Translational Velocity Equations:**

For the fixed point:

$$V_O = 0 \tag{7}$$

For the first moving point (A), the motion is given as

$$V_A = r\omega_\theta \tag{8}$$

For subsequent points, the velocity propagates through linkages

$$V_B = V_A + V_{B/A} \tag{9}$$

For point C, the velocity equation will be

$$V_C = V_B + V_{C/B} \tag{10}$$

Similarly, for remaining points we can get velocity equations.

**Acceleration Equations**

Acceleration is split into normal ($A_x^N$) and tangential ($A_x^t$) components

$$A = A^N + A^t \tag{11}$$

For the fixed point:

$$A_O = 0 \tag{12}$$

For the first moving point ($A$):

$$A_A^N = \frac{(V_A)^2}{r} \tag{13}$$

Assuming constant velocity:

$$A_A^t = 0 \tag{14}$$

For the second moving point ($B$):

$$A_B = A_A + A_{B/A} \tag{15}$$

For point ($C$) we get,

$$A_C = A_B + A_{C/B} \tag{16}$$

Similarly, for all other moving points we can get acceleration equations.

**General Equations for N-DOF System**

$$V_n = V_{n-1} + V_{n/\,n-1} \tag{17}$$

$$A_n = A_{n-1} + A_{n/n-1} \qquad (18)$$

The description of all parameters and symbols are provided below:

| Symbol | Definition | Symbol | Definition |
|---|---|---|---|
| $O$ | Fixed reference point in the structure | $\omega$ | Angular velocity |
| $A, B, C, D, \ldots Q$ | Moving points in the system | $r$ | Radius of the rotating link |
| $V_n$ | Velocity of the n-th point | $V_{n/n-1}$ | Relative velocity of point n with respect to $n-1$ |
| $A_n$ | Acceleration of the n-th point | $A_{n/n-1}$ | Relative acceleration of point n with respect to $n-1$ |
| $A^N$ | Normal acceleration component | $A^t$ | Tangential acceleration component |

**Validation of analytical modelling using GA-SQP-NN and SolidWorks**

a) **Technique for validation with GA-SQP-NN:**

A Genetic Algorithm (GA) starts with conjectures and then develops to improve them. Typically, a GA is divided into five sections, which are:

1. Chromosome: In GAs, solutions are often encoded as a string of chromosomes. Every gene in the chromosome could represent a fraction of the solution.
2. Initial Population: This is the starting set of chromosomes-population. Such a diverse starting population leaves much more hope for finding an optimal solution.
3. The fitness function in this context computes how close the given chromosome is to the solution that one optimally desires. Fitness of chromosomes plays the most important role during the drive of evolution as fitter chromosomes are more likely to reproduce.
4. Selection: This function takes chromosome candidates into the reproductive processes because of their fitness. Some of the common methods of selection are: tournament selection, roulette wheel selection, and rank selection.
5. Crossover and Mutation Operators: These are the genetic operators which generate a new chromosome from the population. The crossover combines a bit of two parent chromosomes to form offspring while mutation introduces random variations to uphold the generation's diversity so that it doesn't converge too early.

A chromosome in a GA is a data structure that can be variable in complexity from as simple as a binary string to very complex data arrangements depending on the problem that is to be solved. The population of chromosomes can be randomly generated or even manually taken up, hence providing diverse starting points for the algorithm. The fitness function shows how good each chromosome is at performing the task at hand compared to the desired goal or solution and thus guides the selection process. The mechanism of selection selects which chromosomes most probably reach the next generation after possibly undergoing genetic operations such as mutation and crossover. It is facilitated by the crossover operator, which allows the genetic material to be interchanged between two chromosomes, hence creating two offspring inheriting traits from both parents. This will enhance diversity and enable the exploration of varying solutions. It introduces randomness in the algorithm through mutation operators, which modify a portion or set of genes of a chromosome. Through this mechanism, the algorithm can see new possibilities and not get trapped in local optima. Hence, all these processes drive the evolutionary improvement of solutions in summary. GAs have a wide range of applications in a variety of areas include credit scoring model [20], heart disease diagnosis [21], image reconstruction [22], forecasting of crude oil price model [23], optical metasurfaces model [24], and mobile robot model for path planning [25].

This algorithm is a fusion between Genetic algorithm (GA) and Sequential Quadratic Programming (SQP) in order to optimize the weights of a Neural Network (NN). Table 5 is a step-by-step structured breakdown and description of how this algorithm can implemented for our scenario:

**Table 5:** GA-SQP-NN optimization process pseudocode for Kinematic analysis

> **Input:** The FNN decision parameters are represented by the chromosome with identical genes as:
> 
> "$W = [W_{L_V}, W_{A_V}, W_{L_A}, W_{A_A}]$" for $W_{L_V} = [\phi_{L_V}, \eta_{L_V}, b_{L_V}]$, $W_{A_V} = [\phi_{A_V}, \eta_{A_V}, b_{A_V}]$, $W_{L_A} = [\phi_{L_A}, \eta_{L_A}, b_{L_A}]$, $W_{A_A} = [\phi_{A_A}, \eta_{A_A}, b_{A_A}]$.
> 
> **Population:** A population is defined by the number of chromosomes in a collection:
> 
> $W = [W_1, W_2, W_3, ..., W_k]^T$ for n-th component
> 
> **Output:** $W_{GA-Best}$ represents the optimal GA-SQP-NN global weights.

**Initialization:** Use pseudo-random numbers to create chromosomes. Using the commands "GA" and "gaoptimset" together with the appropriate settings and explanations, the starting process is executed.

**Fitness Assessment:** For each "$W$" in $P$, adjust fitness $\hat{e}$ and correspondingly for all four scenarios

**Closure:** If any of the following conditions are fulfilled, end the program:
- "Obtained fitness": $e=10^{-20}$
- "Population size": 30*10, StallGenLimit→100
- "Tolerances": [TolCon=$10^{-20}$ & TolFun=$10^{-18}$], Generation=20
- Remaining values are set as default GA routine

Move toward **Storage** step

**Rank:** Each "$W$" in $P$ has a rank that indicates the values of fitness that have been achieved.

**Reproduction:**
- Selection
- Mutation
- Crossover
- Elitism

Go to **fitness evaluation** step

**Storage:** Save $W_{GA-Best}$, fitness evaluation $\hat{e}$, time, generations and function counts for current run of GAs.

**End of GA process**

**Start of SQP process:**

**Input:** $W_{GA-Best}$ is the starting point

**Output:** $W_{GA-SQP}$ indicates the best weights in SQP.

**Initialization:** Set the number of iterations, the number of constraints, and other parameters for "optimset".

**Closure:** When any of the following criteria meets, terminate procedure:
- "Fitness" $e=10^{-18}$
- "Tolerances": [TolCon=TolFun=TolX=$10^{-22}$], Generation=1000
- "MaxFunEvals"=200,000

While (Stop when criteria meet)

> **Fitness Evaluation:** Adjust fitness for every "$W$" in $P$.
>
> **Fine-tuning:** Adjust "fmincon" using the SQP approach to tune $W$.
>
> **Accumulate:** Save $W_{GA-SQP}$, fitness evaluation $\hat{e}$, time, generations, counts of function for multiple run of SQP.
>
> **End of SQP process**
>
> **Data Generations:** Use GA-SQP to repeat the procedure 100 times in order to obtain a bigger dataset of ANN optimization variables that will allow for analytical computation of the nonlinear epidemiological smoking model.

(a) Linear Velocity

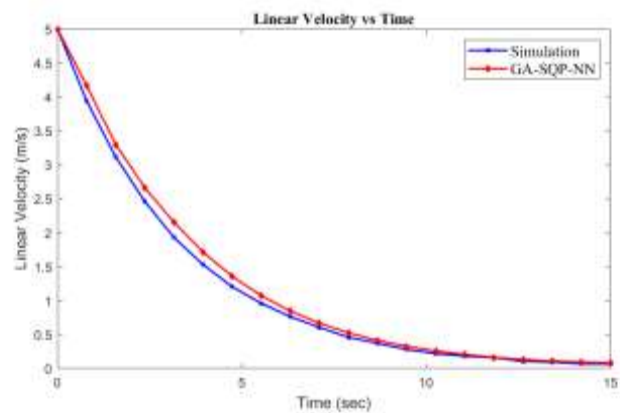

(b) Angular Velocity

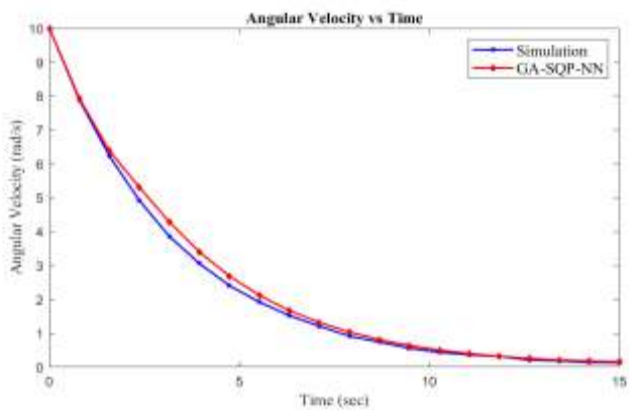

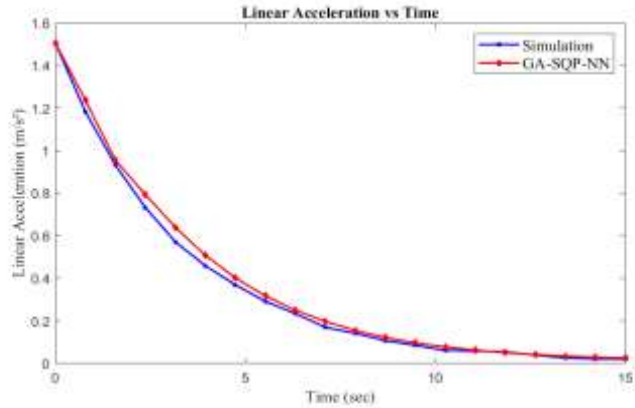

(c) Linear Acceleration

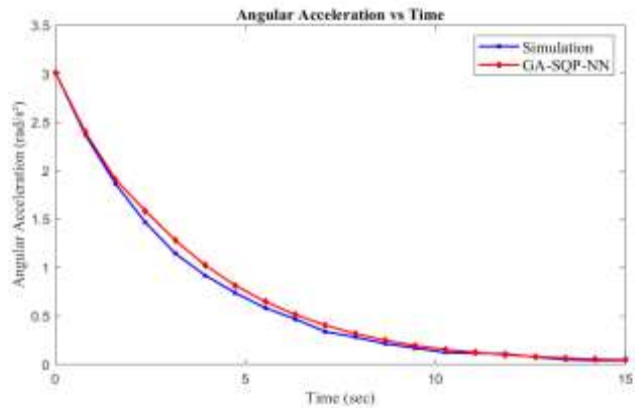

(d) Angular Acceleration

**Fig. 7** Kinematic Analysis using SolidWorks and Genetic algorithm with sequential quadratic programming and neural network

Figure 7 illustrates the kinematic analysis of TSDTM, comparing SolidWorks simulations with optimization-based approaches. The graphs show a smooth decline in linear and angular velocity, as well as acceleration. This indicates a stable deployment. The close match between simulation and optimization results confirms the accuracy and efficiency of the mechanism.

**Analytical Dynamics**

- The deployable structure consists of 12 identical units forming a circular structure.
- Each unit has mass m, and they all move as the structure stows or deploys.
- The motion causes a change in radius $R$ and height $L$ of each unit.
- We describe the motion using an angular displacement $\theta$.
- We used small angle approximations i.e. $sin\theta \approx \theta$ and $cos\theta \approx \theta$.

Each unit has two types of motion contributing to kinetic energy:

1. Radial motion (change in $R$)
2. Vertical motion (change in $L$)

From geometry:

- The radial displacement of each unit is proportional to $R_0 \theta$.
- The vertical displacement is proportional to $L\theta$.

$$v_R = R_0 \dot{\theta}$$
$$v_L = L\dot{\theta}$$

As,
$$v^2 = v_R^2 + v_L^2$$
$$v^2 = (R_0\dot{\theta})^2 + (L\dot{\theta})^2$$

**Kinetic Energy:**

Kinetic energy of a single unit is given by:

$$T_{unit} = \frac{1}{2}mv^2$$

$$T_{unit} = \frac{1}{2}m(R_0^2\dot{\theta}^2 + L^2\dot{\theta}^2)$$

$$T_{unit} = \frac{1}{2}m(R_0^2 + L^2)\dot{\theta}^2$$

Total kinetic energy of the structure (12 units):

$$T = \frac{1}{2}mv^2 + \frac{1}{2}mv^2 + \frac{1}{2}mv^2 + \frac{1}{2}mv^2 + \frac{1}{2}mv^2 + \frac{1}{2}mv^2 + \frac{1}{2}mv^2 + \frac{1}{2}mv^2 + \frac{1}{2}mv^2$$
$$+ \frac{1}{2}mv^2 + \frac{1}{2}mv^2 + \frac{1}{2}mv^2$$

$$T = 12 \times \frac{1}{2}mv^2$$

$$T = 6m(R_0^2 + L^2)\dot{\theta}^2$$

**Potential Energy:**

Potential energy is given by:

$$V = V_e + V_g$$

Where,
- $V_e$ is the elastic potential energy.
- $V_g$ is the gravitational potential energy.

**Elastic Potential Energy:**

$$V_{unit,e} = \frac{1}{2}kx^2$$

As, $x = R_0\theta$

$$V_{unit.e} = \frac{1}{2}k(R_0\theta)^2$$

Elastic potential energy of all units:

$$V_e = 12 \times \frac{1}{2}k(R_0\theta)^2$$

$$V_e = 6kR_0^2\theta^2$$

**Gravitational Potential Energy:**

$$V_{unit,g} = mgh$$

As, $h = L\theta$

$$V_{unit,g} = mgL\theta$$

Gravitational potential energy of all units:

$$V_g = 12mgL\theta$$

**Total Potential Energy:**

Total potential energy of complete structure is given by:

$$V = 6kR_0^2\theta^2 + 12mgL\theta$$

**Energy Method:**

Now,

$$T + V = 6m(R_0^2 + L^2)\dot\theta^2 + 6kR_0^2\theta^2 + 12mgL\theta$$

Taking derivative and putting it equal to 0.

$$\frac{d}{dt}(T + V) = \frac{d}{dt}(6m(R_0^2 + L^2)\dot\theta^2 + 6kR_0^2\theta^2 + 12mgL\theta)$$

$$0 = 6m(R_0^2 + L^2)(2\dot\theta\ddot\theta) + 6kR_0^2(2\theta\dot\theta) + 12mgL\dot\theta$$

$$0 = (12m(R_0^2 + L^2)\ddot\theta + 12kR_0^2\theta + 12mgL)\dot\theta$$

Since, $\dot\theta \neq 0$

$$12m(R_0^2 + L^2)\ddot\theta + 12kR_0^2\theta + 12mgL = 0$$

$$\ddot\theta + \frac{12kR_0^2}{12m(R_0^2 + L^2)}\theta + \frac{12mgL}{12m(R_0^2 + L^2)} = 0$$

$$\ddot{\theta} + \frac{kR_0^2}{m(R_0^2 + L^2)}\theta + \frac{gL}{(R_0^2 + L^2)} = 0$$

Neglect gravity term for natural frequency. And compare with

$$\ddot{\theta} + \omega_n^2 \theta = 0$$

$$\omega_n = \sqrt{\frac{kR_0^2}{m(R_0^2 + L^2)}}$$

Substituting $L = 6.47$, $k = 1$, $m = 1$, and $R_0 = 12.5$, we get

$$\omega_n = 0.888 \; rad/s$$

$$f_n = \frac{\omega_n}{2\pi}$$

$$f_n = 0.1414 \; Hz$$

In ANSYS, we conducted simulation of dynamic analysis across multiple antenna structures, in order to determine their natural frequencies. The simulation results are illustrated in Fig. 8 and Fig. 9 for with and without links antenna on different (6m, 15m, 25m and 30m) apertures configuration.

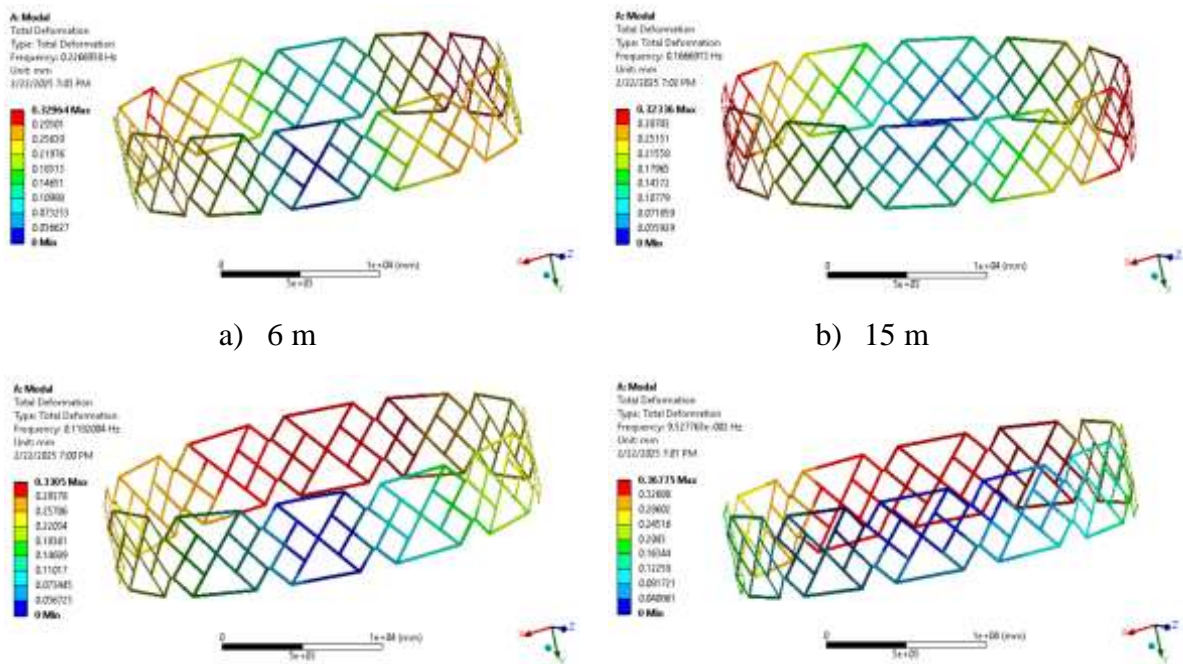

a) 6 m                                   b) 15 m

c)  25 m                              d)  30 m

**Fig. 8** Dynamic analysis of TSDTM with links for different apertures

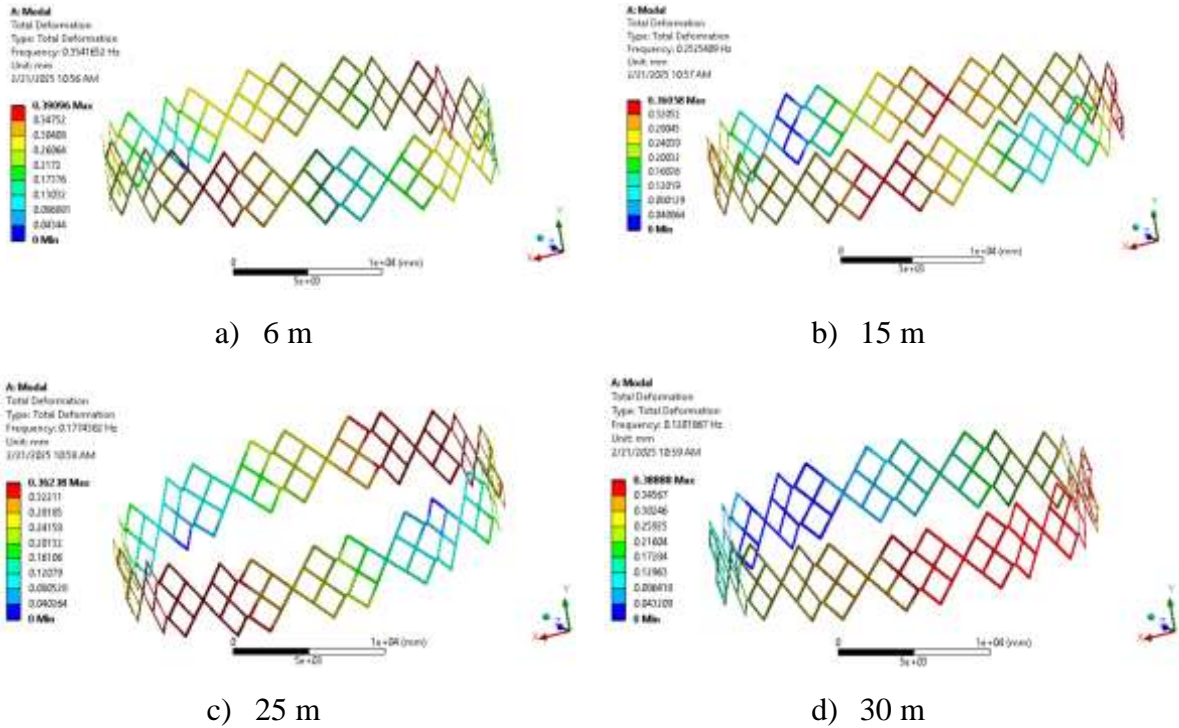

a)  6 m                               b)  15 m

c)  25 m                              d)  30 m

**Fig. 9** Dynamic analysis of TSDTM without links for different apertures

The table provides a comparison between the natural frequency and frequency response of the simulation of the deployable space truss mechanism as a function of mass and aperture size. In both configurations, with the increase in the size of the aperture, the mass increases and, therefore, causes the natural frequency to decrease correspondingly. Notably, the "without links" configuration has higher frequency values compared to the "with links" configuration, that is, a stiffer yet vibrationally more sensitive structure. For instance, at an aperture of 6m, the frequency is 0.2266 Hz (with links) and 0.35417 Hz (without links), and at 30m, it decreases to 0.09527 Hz (with links) and 0.12019 Hz (without links). The coupled frame, although potentially to provide strength, has some extra mass to affect its dynamic response. The uncoupled frame gives more stiffness and elevated frequency, and the coupled structure gives lower frequency response due to extra mass and potential flexural activities.

| Aperture | Natural Frequency | Storage Ratio | Simulation Frequency Responses ||
|---|---|---|---|---|
| | | | With Links | Without Links |
| **6m** | 1.2 Hz | - | 0.2266 | 0.3541 |
| **6m(2)** | 1.0 Hz | - | - | - |
| **13m** | 0.14 Hz (14 units) <br> 0.25 Hz (7 units) | - | - | - |
| **15m** | 0.85 - 1 Hz | - | 0.1666 | 0.2525 |
| **25m** | 0.0652 Hz | 17.14 | 0.1182 | 0.1774 |
| **28m** | 0.09 Hz | - | - | - |
| **30m** | 0.0824 Hz | - | 0.09527 | 0.1201 |

**Natural Frequency of Existing and Currently Studied Antennas**

| Antenna Configuration | Natural Frequency (Hz) |
|---|---|
| AstroMesh. | 0.012 |
| H-double | 0.021 |
| Single ring deployable truss | 0.090 |
| Double ring deployable truss | 0.101 |
| **Triple scissors truss mechanism (Novel Design)** | **0.1182** |
| Double scissor link deployable mechanism | 0.0652 |

Resonance is a critical factor in determining the structural stability and performance. When natural frequency aligns with external excitations, excessive vibrations can occur which can lead to structural failure. As the requirements of this antenna are highly precise due to space applications, it is important to analyze the natural frequencies to ensure they remain above the range of operational and environmental disturbances such as micro-vibrations from spacecraft mechanics and thermal fluctuations.

Typical spacecraft-induced disturbances range from 0.01 Hz to 10 Hz, meaning the antenna's natural frequency should be designed well above this range to avoid resonance. Advanced damping techniques and structural optimization, such as increasing hinge stiffness or implementing tuned mass dampers, can reduce resonance effects.

**Material and Geometry Optimization**

To meet the stringent weight and strength requirements of space missions, this section focuses on material selection and geometry optimization. A Support Vector Machine (SVM) is used to identify suitable materials under LEO thermal conditions, while a Neural Network coupled with gradient-based methods is employed to refine geo- metric parameters. The optimized results demonstrate improved frequency response, reduced mass, and enhanced deformation resistance, confirming the potential of AI- based optimization in deployable structural design.

*Material Selection and Optimization*

In the structural design of aerospace structures, especially those for Low Earth Orbit (LEO) applications, material selection is an important choice with a direct impact on performance, safety, and mission life. Materials need to endure the extreme ther- mal, mechanical, and environmental conditions of space, including extreme variations of temperature and exposure to radiation. In order to systematically determine ap- propriate materials, a data-driven strategy was utilized based on a combination of mechanical properties (tensile strength, Young's modulus, and density) and thermal limits (upper and lower operating temperatures). The process of selection commenced by creating a dataset of prospective materials, with each material characterized by key performance indicators. In order to identify suitability under LEO environments, materials were initially filtered according to thermal survivability, that is, a mini- mum of -100°C and a maximum requirement of at least 150°C. Such limits are the typical operating extremes found in space. In order to augment decision-making, a supervised machine learning method, Support Vector Machine (SVM), was learned on normalized mechanical features for the purpose of determining whether a material was suitable or unsuitable. Of the materials categorized as suitable, an optimization scoring function was used to determine the optimal material. The scoring function favored high tensile strength and modulus and low density to combine both structural strength and mass efficiency. This process not only guarantees thermal compatibility but also maximizes mechanical performance, offering a strong and efficient method- ology for material selection for space applications. The details of the materials are given table below.

**Table. Materials details**

| Material | Young's Modulus | Density ($g/cm^3$) | Poisson's Ratio | CTE ($\mu m/m.°C$) | Yield Strength (MPa) | Tensile Strength | Ultimate Strength | Elastic Limit (MPa) | Breaking Strength | Ductile / Brittle | Max Temperature (°C) |
|---|---|---|---|---|---|---|---|---|---|---|---|
| T1100G | 320 | 1.8 | 0.29 | 0.3 | 2350 | 3050 | 3350 | 2150 | 3250 | B. | 140 |

| | | | | | | | | | | |
|---|---|---|---|---|---|---|---|---|---|---|
| CFRP | | | | | | | | | | |
| M55J/954-6 | 290 | 1.88 | 0.3 | 0.2 | 1580 | 1950 | 2020 | 1480 | 1980 | B. | 150 |
| Cyanate Ester | 160 | 1.46 | 0.32 | 0.35 | 1050 | 1420 | 1550 | 980 | 1450 | B. | 180 |
| Ti-6Al-4V | 110 | 4.42 | 0.31 | 8.6 | 860 | 930 | 940 | 850 | 920 | D. | 400 |
| Al-7075-T7351 | 70 | 2.8 | 0.33 | 21 | 490 | 550 | 560 | 480 | 540 | D. | 120 |
| FeNi36 Alloy | 140 | 8.05 | 0.29 | 1.2 | 230 | 480 | 490 | 220 | 470 | D. | 300 |
| Be-S-65 Grade | 285 | 1.84 | 0.03 | 11.3 | 230 | 410 | 440 | 220 | 420 | B. | 200 |

Through the use of the material selection and optimization algorithm outlined above, the list of candidate materials was tested against both thermal performance and mechanical performance specifications. Once filtered for applicability under Low Earth Orbit (LEO) conditions—namely, materials that can sustain temperatures above 150°C and below -100°C—only a limited number of materials were able to meet the necessary thermal thresholds. A Support Vector Machine (SVM) classifier was trained on the normalized mechanical properties to classify material appropri- ateness, and the model validated these thermally qualified candidates. To define the best material, an optimization score for each appropriate material was calculated with regard to tensile strength, Young's modulus, and density, focusing on maximiz- ing the mechanical performance while minimizing weight. Among all the materials, Carbon Fiber Reinforced Polymer (CFRP) was found to be the best material due to its strength-to-weight ratio and stiffness. To further confirm this choice, a deformation analysis presented in Figure below was conducted under simulated LEO loading conditions. The analysis showed that CFRP had the minimum deformation among all materials, proving its highest structural integrity and dimensional stability in space. This verification emphasizes CFRP's ability to be used in aerospace applications in which mechanical performance and thermal resistance are paramount.

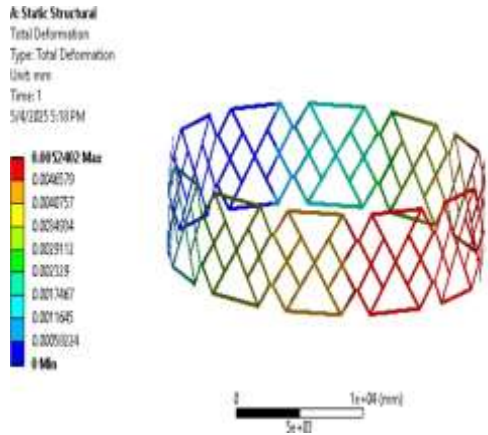
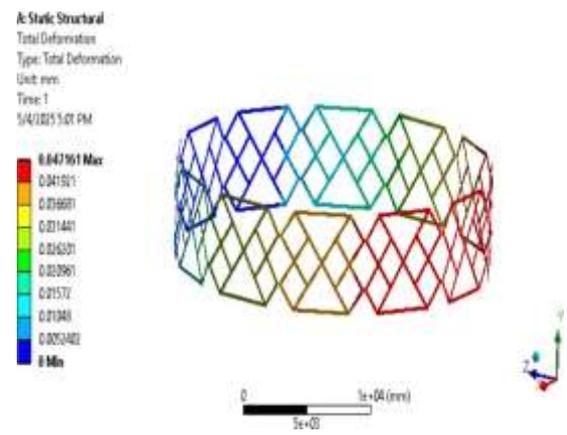

(a) M55J/954-6 CFRP  (b) T1100G CFRP

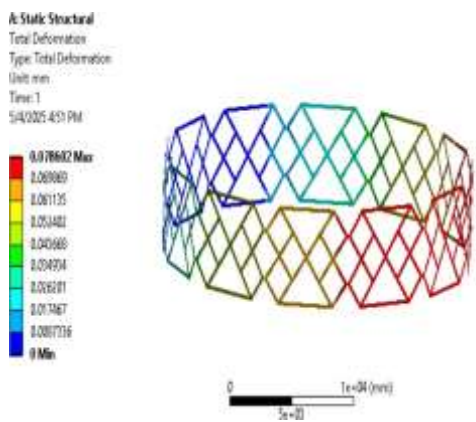
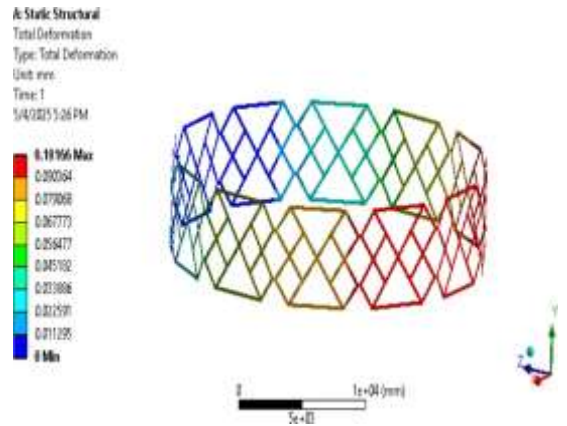

(c) Cyanate Ester CFRP  (d) C/SiC

**Figure.** Deformation Analysis for different Materials

**Geometry Optimization**

Machine learning (ML) based geometry optimization of truss structures is a state of the art method that combines computational intelligence and structural engineering to obtain efficient, lightweight, and durable structures. Geometric optimization of trusses classically means iterative numerical approaches to optimize material use while meeting strength, stability, and deformation requirements. These procedures can be costly in terms of computation, particularly for large and intricate truss systems. Machine learning presents a strong alternative through learning from past design experience data or simulation output to forecast best geometries more effectively. In deployable space antenna design for large apertures, structural geometry is central to dynamic performance and deployment efficiency determination. One of the main challenges is to optimize the geometry specifically the radius and individual link lengths such that it becomes dynamically stable and meets stringent mass and frequency requirements. This work introduces a gradient-based optimization method through Sequential Quadratic Programming (SQP) to reduce the natural frequency of the deployable antenna such that the radius is greater than a predetermined limit and the generated frequency is within manageable operating limits. Through the incorporation of geometric parameters, contributions to the inertia, and mechanical limitations into a single optimization platform, the technique offers a systematic approach to the trade-off of stiffness, mass distribution, and deployability. The importance of this method is that it can produce light weight, low frequency structures to deploy space without sacrificing structural integrity or accuracy.

The optimized radius is 13.65m, optimized frequency 0.1107Hz and optimized lengths which are presented in Table.

**Table** Original and Optimized Results

| Link | Diameter | $L_1$ | $L_2$ | $L_3$ | $L_4$ | $L_5$ | $L_6$ | $L_7$ |
|---|---|---|---|---|---|---|---|---|
| Original | 25 | 6.64 | 6.64 | 2.14 | 2.14 | 2.14 | 2.14 | 3.32 |
| Optimized | 27.3 | 7.09 | 7.09 | 2.41 | 2.41 | 2.41 | 2.41 | 3.54 |
| Link | $L_8$ | $L_9$ | $L_{10}$ | $L_{11}$ | $L_{12}$ | $L_{13}$ | $L_{14}$ | |
| Original | 3.32 | 3.32 | 3.32 | 1.66 | 1.66 | 1.66 | 1.66 | |
| Optimized | 3.54 | 3.54 | 3.54 | 1.77 | 1.77 | 1.77 | 1.77 | |

The optimized single deployed modular unit of optimized TSDAM with dimensions is presented in Figure.

Figure. Single Modular Unit of Optimized Antenna

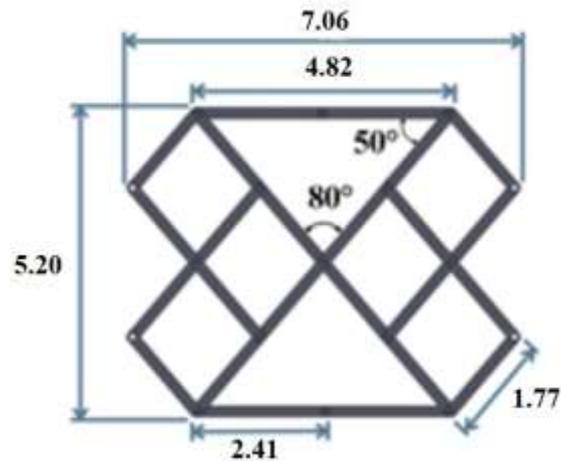

**Dynamic Analysis of Optimized TSDTM**

In ANSYS, we performed simulation of dynamic analysis across antenna structure of a 27.3 m aperture, to determine their natural frequency. The simulation results are illustrated in Figure 9.5 for optimized TSDTM antenna. The modal analysis reveals a natural frequency of 0.10859 Hz for the structure.

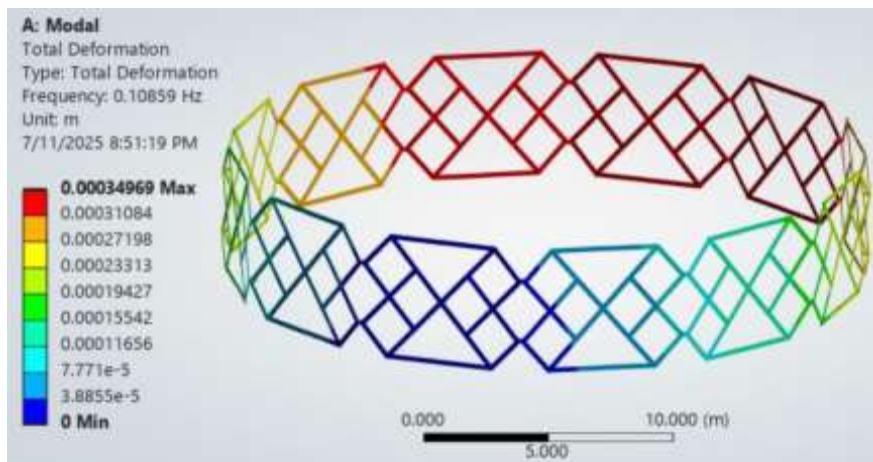

Figure. Dynamic analysis of optimized TSDTM

*Comparison of Optimized Frequency and Simulation*
The modal analysis calculated the natural frequency of the structure as 0.10859 Hz, while the machine learning optimized frequency was 0.1107 Hz. The absolute difference between the two frequencies is 0.00211 Hz, and the relative difference is about 1.94%. This minor difference infers the high degree of accuracy and credibility of the machine learning model in the prediction of structural dynamic behavior, with excellent closeness to the results of traditional simulation methodologies.

**Conclusion:**

This paper presents the TSDTM as an advanced solution for large-aperture deployable antennas. First through kinematic and dynamic analyses, validated by simulations in MSC ADAMS and ANSYS, it is proven beyond any reasonable doubt that the mechanism is far more compact, structurally stable, and deployable than its counterparts. Considered results show that it has high natural frequency, which would guarantee it to resist resonant vibrations when applied in space. Further work, therefore, will be directed at enhancing performance optimization leading to real-world structural deployment actuation reliability improvements.